\documentclass[10pt,twocolumn,letterpaper]{article}

\usepackage{wacv}
\usepackage{times}
\usepackage{epsfig}
\usepackage{graphicx}
\usepackage{amsmath}
\usepackage{amssymb}
\usepackage{multirow}
\usepackage{booktabs}
\usepackage{array}
\usepackage[colorlinks,hyperindex,breaklinks]{hyperref}
\DeclareUnicodeCharacter{3000}{ }

\def\wacvPaperID{680} % Enter the WACV Paper ID here 680

\wacvfinalcopy % *** Uncomment this line for the final submission
\ifwacvfinal
\def\assignedStartPage{1} % *** Enter the assigned starting page number (instead of 9876)
\fi

\ifwacvfinal
\else
\fi

\begin{document}

%%%%%%%%% TITLE
\title{Defect-GAN: High-Fidelity Defect Synthesis for Automated Defect Inspection}

\author{
Gongjie Zhang$^1$ \qquad Kaiwen Cui$^1$ \qquad Tzu-Yi Hung$^2$ \qquad Shijian Lu\thanks{Corresponding author.} $^{1}$ \smallskip\\
{$^{1}$Nanyang Technological University \qquad \quad $^{2}$Delta Research Center, Singapore}\\
{\tt\small \{gongjiezhang,shijian.lu\}@ntu.edu.sg \quad kaiwen001@e.ntu.edu.sg \quad tzuyi.hung@deltaww.com}
}

\maketitle
%\thispagestyle{empty}

%%%%%%%%% ABSTRACT
\begin{abstract}
Automated defect inspection is critical for effective and efficient maintenance, repair, and operations in advanced manufacturing. On the other hand, automated defect inspection is often constrained by the lack of defect samples, especially when we adopt deep neural networks for this task. This paper presents Defect-GAN, an automated defect synthesis network that generates realistic and diverse defect samples for training accurate and robust defect inspection networks. Defect-GAN learns through defacement and restoration processes, where the defacement generates defects on normal surface images while the restoration removes defects to generate normal images. It employs a novel compositional layer-based architecture for generating realistic defects within various image backgrounds with different textures and appearances. It can also mimic the stochastic variations of defects and offer flexible control over the locations and categories of the generated defects within the image background. Extensive experiments show that Defect-GAN is capable of synthesizing various defects with superior diversity and fidelity. In addition, the synthesized defect samples demonstrate their effectiveness in training better defect inspection networks.
\end{abstract}

%%%%%%%%% BODY TEXT
\section{Introduction}

\begin{figure}[t!] 
\begin{center}
   \includegraphics[width=0.97\linewidth, height=76.75mm]{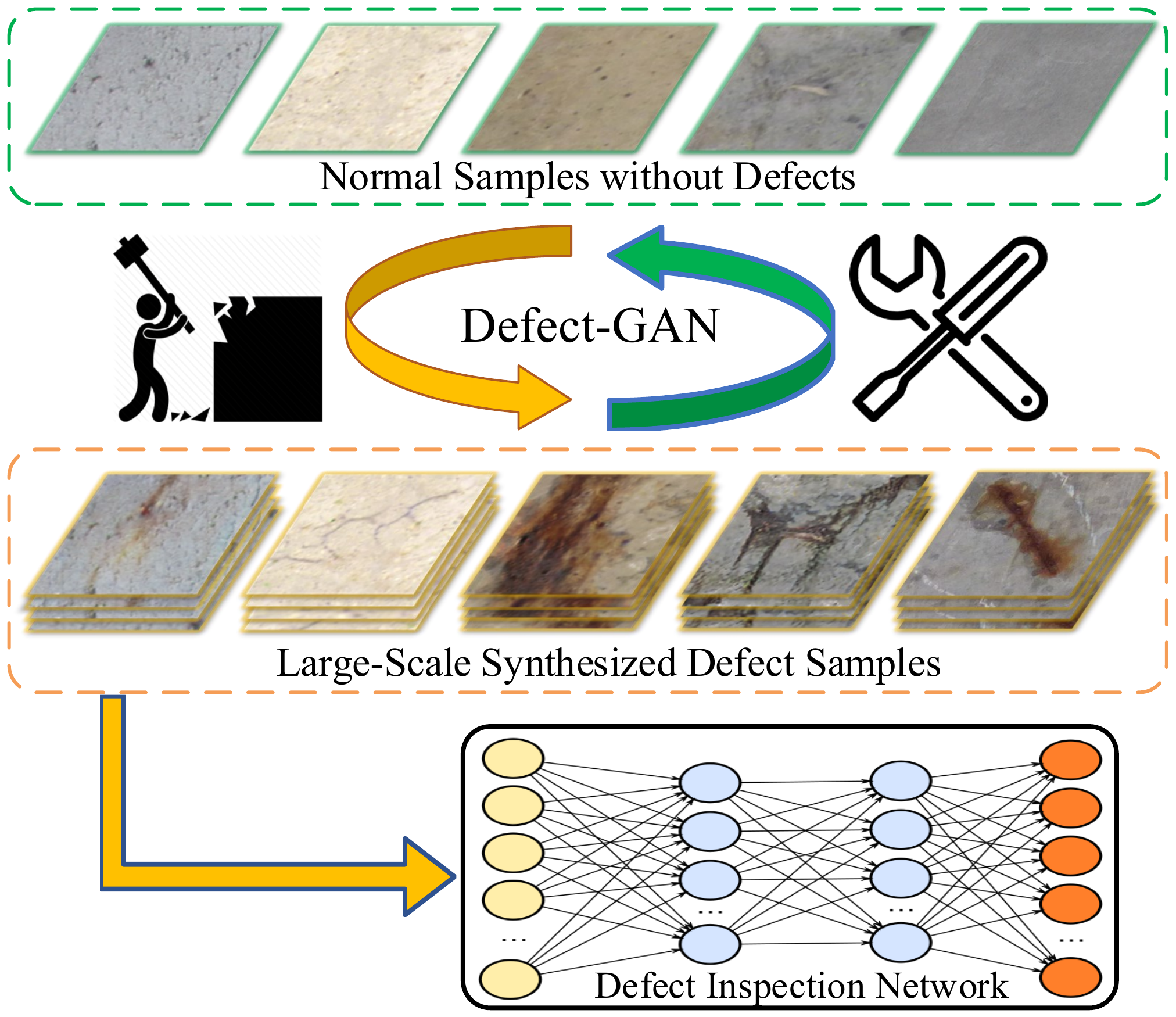}
\end{center}
   \caption{Mimicking the defacement and restoration processes over the easily collected normal samples, Defect-GAN generates large-scale defect samples with superior fidelity and diversity. The generated defect samples demonstrate great effectiveness in training accurate and robust defect inspection network models.
   }
\label{fig:fig1}
\end{figure}

Automated visual defect inspection aims to automatically detect and recognize various image defects, which is highly demanded in different industrial sectors, such as manufacturing and construction. In manufacturing, it is one key component in maintenance, repair, and operations (MRO) that aims to minimize the machinery breakdown and maximize production. It is also important for quality control for spotting anomalies at different stages of the production pipeline. In construction, it is critical to public safety by identifying potential dangers in various infrastructures such as buildings, bridges, etc. Although automated visual defect inspection has been studied for years, it remains a challenging task with a number of open research problems.

One key challenge in automated visual defect inspection lies with the training data, which usually manifests in two different manners. First, collecting a large number of labeled defect samples are often expensive and time-consuming. The situations become much worse due to the poor reusability and transferability of defect samples, i.e., we often have to re-collect and re-label defect samples while dealing with various new defect inspect tasks. Second, collecting defect samples is not just about efforts and costs. In many situations, the defect samples are simply rare, and the amount available is far from what is required, especially when training deep neural network models. The availability of large-scale defect samples has become one bottleneck for effective and efficient design and development of various automated defect inspection systems.

An intuitive way to mitigate the defect insufficiency issue is to synthesize defect samples. Though Generative Adversarial Networks (GANs) have achieved superior image synthesis in recent years, synthesizing defect samples using GANs is still facing several challenges. \textit{First}, existing GANs usually require large-scale training data, but large-scale defect samples are not available in many situations. \textit{Second}, GANs tend to generate simpler structures and patterns by nature \cite{seeing_gan} and so are not good at synthesizing defects that often have complex and irregular patterns with large stochastic variations. \textit{Third}, defect samples with different backgrounds are very difficult to collect, and GANs thus tend to generate defect samples with similar backgrounds as the collected reference samples. As a result, the GANs synthesized defect samples often have similar feature representation and distribution as those reference samples and offer little help while facing various new defect samples on various different backgrounds.

Inspired by~\cite{mery2002automated} that collects defect samples by manually damaging the surface of normal work-pieces, we design a Defect-GAN that aims for automated generation of high-fidelity defect samples for training accurate and robust defect inspection networks. Defect-GAN simulates the defacement and restoration processes, which greatly mitigates the defect-insufficiency constraint by leveraging large-scale normal samples that are often readily available. We design novel control mechanisms that enable Defect-GAN to generate different types of defects at different locations of background images flexibly and realistically. We also introduce randomness to the defacement process to capture the stochastic variation of defects, which improves the diversity of the generated defect samples significantly. Additionally, we design a compositional layer-based network architecture that allows for generating defects over different normal samples but with minimal change of normal samples' background appearance. As a result, the model trained with such generated defect samples is more capable of handling new defect samples with variously different backgrounds. Extensive experiments show that Defect-GAN can generate large-scale defect samples with superior fidelity and diversity as well as effectiveness while applied to train deep defect inspection networks.

\smallskip 
The contributions of this work can be summarized in three aspects. \textit{First}, we design a compositional layer-based network architecture to generate defects from normal samples while preserving the appearance of normal samples, which improves the defect diversity by simulating how defects look like on various normal samples. \textit{Second}, we propose a Defect-GAN that synthesizes defects by simulating defacement and restoration processes. It offers superior flexibility and control over the category and spatial locations of the generated defects in the image background, achieves great defect diversity by introducing stochastic variations into the generation process, and is capable of generating high-fidelity defects via defacement and restoration of normal samples. \textit{Third}, extensive experiments show that the Defect-GAN generated defect samples help to train more accurate defect inspection networks effectively.

\begin{figure*}[t!] 
\begin{center}
   \includegraphics[width=1.0\linewidth, height=66mm]{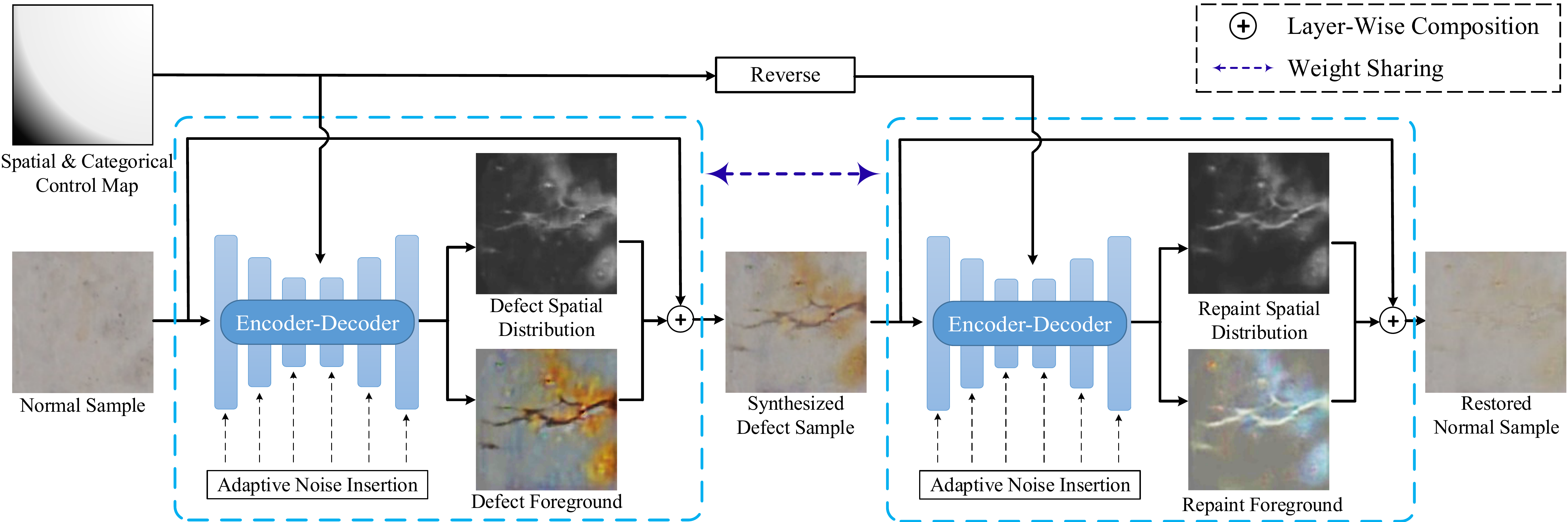}
\end{center}
   \caption{Generation pipeline of the proposed Defect-GAN: It adopts an encoder-decoder structure to synthesize defects by mimicking defacement and restoration processes. The \textit{Spatial \& Categorical Control Map} generated from category vectors controls where and what kind of defects to generate within the provided normal sample. The \textit{Adaptive Noise Injection} introduces stochastic variations into the generated defects to improve the diversity of the generated defects. In addition, Defect-GAN adopts a \textit{Layer-Wise Composition} strategy that produces defect and repaint foregrounds according to the corresponding spatial distribution maps. This helps preserve the style and appearance of the normal samples and achieve superior realism in defect synthesis.
   }
\label{fig:model}
\end{figure*}

\section{Related Works}

\noindent {\bf Image Synthesis.}
GANs~\cite{gan} are a powerful generative model that simultaneously trains a generator to produce realistic faked images and a discriminator to distinguish between real and faked images. Early attempts~\cite{gan,dcgan,wgan,progressivegan,biggan} focus on synthesizing images unconditionally. Recently, more and more works emerge to perform image synthesis conditioned on input images, which has wide applications including style translation~\cite{unit,pix2pix,discogan,CycleGAN,AttnCycleGAN,koksal2020RF-GAN}, facial expression editing~\cite{StarGAN,ganimation,chen2019semantic,CascadeEFGAN,leed}, super-resolution~\cite{ledig2017photo,wang2018cascaded,sajjadi2017enhancenet}, image inpainting~\cite{yeh2017semantic,yu2018generative,pathak2016context,yan2018shift}, etc. Another trend is multi-modal image synthesis~\cite{StyleGAN,StyleGANv2,munit,StarGANv2,smis}. 
However, existing methods fail to generalize well on defect synthesis. Our Defect-GAN is designed to generate defect samples by simulating the defacement and restoration processes and incorporating randomness to mimic the stochastic variations within defects. Besides, inspired by~\cite{lrgan,finegan,PizzaGAN,zhan2019spatial}, it deems defects as a special foreground and adopts a layer-based architecture to compose defects on normal samples, thus reserve the normal samples' style and appearance and achieving superior synthesis realism and diversity.

\smallskip
\noindent {\bf Learning From Limited Data.}
Deep learning based techniques~\cite{FasterRCNN,cadnet,Zhan_2018_ECCV} usually require a large amount of annotated training samples, which are not always available. Recent researches have proposed many attempts to mitigate the data-insufficiency issue. They can be broadly categorized as few-shot learning and data augmentation.

Few-shot learning~\cite{ProtoNet,RelationNetwork,DynamicFewshotWOForgetting,CloserFewshotClassification,MetaOptNet,FewshotReweighting,metarcnn,metadet,Zhang_2021_WACV,fsod,fsdet,Zhang2021MetaDETRFO} refers to learning from extremely limited training samples (e.g., 1 or 3) for an unseen class. However, their performances are quite limited and thus far from practical application. Besides, few-shot learning techniques usually require large amounts of samples from the same domain, which does not lift the data-insufficiency constraint. 
Data augmentation aims to enrich the training datasets in terms of quantity and diversity such that better deep learning models can be trained. 
Several recent research attempts~\cite{dagan,lowshotimaginary,fvaegand2,SDGAN} adopt GANs as data augmentation methods to synthesis realistic training samples. 
The proposed Defect-GAN also works as a data augmentation method to train better defect inspection networks by synthesizing various defect samples with superior diversity and fidelity.

\smallskip
\noindent {\bf Defect Inspection.}
Surface defect inspection refers to the process of identifying and localize surface defects based on machine vision, which is an important task with extensive real-life applications in industrial manufacturing, safety inspection, building construction, etc. Before deep learning era, traditional methods~\cite{ngan2011automated,tsai1999automated,kuo2014automatic,zhou2014sparse,tolba2015multiscale} design hand-crafted feature extractors and heuristic pipelines, which needs specialized expertise and not robust. In deep learning era, many works~\cite{li2016deformable,faghih2016deep,defectgan,CODEBRIM} adopt Convolutional Neural Networks (CNNs) based models for defect inspection and achieve remarkable performances.

However, in practical scenarios, limited number of defect samples has always been a bottleneck issue. To mitigate such defect-insufficiency issue, \cite{mery2002automated} manually destroys work-pieces to collect defect samples; \cite{mery2005simulation,huang2009template} further adopt Computer-Aided Drawing (CAD) to synthesis defect samples. However, such methods can only handle simple cases. The recently proposed SDGAN~\cite{SDGAN} adopts GANs to perform defect sample synthesis for data augmentation. 
We also propose to synthesis defect samples with GANs for training better defect inspection networks. 
By simulating the defacement and restoration processes with a layer-wise composition strategy, our proposed Defect-GAN can generate defect samples with superior realism, diversity, and flexibility. 
It can further provide better transferability by imposing learnt defect patterns on unseen surfaces.

\section{Methodology}

In this section, we discuss the proposed method in details. As illustrated in Fig.~\ref{fig:fig1}, our proposed method consists of two parts: (1) Defect-GAN design for automated synthesis of defect samples, and (2) defect inspection by using the synthesized defect samples.

\subsection{Defect-GAN for Defect Synthesis}

We hypothesis that there are sufficient amount of normal samples, and only a limited number of defect samples since defects are usually rare and difficult to capture.
Based on this hypothesis of data availability, we propose to perform defect synthesis following the paradigm of unpaired image-to-image translation~\cite{CycleGAN,StarGAN}, which usually requires less training data and can produce better synthesis fidelity. 
Our proposed Defect-GAN is based on the intuition that defects do not exist out of thin air, i.e., there is always a defacement process to generate defects over those normal samples, and there also exists a restoration process to restore the defect samples back to normal samples. By mimicking the defacement and restoration processes as mentioned above, we are able to leverage the large number of normal samples to generate required defect samples.

The Defect-GAN architecture consists of a generator $G$ and a discriminator $D$. During the training stage, Defect-GAN performs image translation using $G$ in two cycles: $ n \rightarrow d \rightarrow \hat{n}$ and $d \rightarrow n \rightarrow \hat{d} $, where $n \in R^{H \times W \times 3}$ denotes a normal sample, $d \in R^{H \times W \times 3}$ denotes a defect sample, and $\hat{n}, \hat{d} \in R^{H \times W \times 3}$ denote restored normal and defect sample, respectively. Since the two cycles are identical and simultaneously conducted, we only describe the cycle $n \rightarrow d \rightarrow \hat{n} $ in the following sections for simplicity.

The generator $G$ is illustrated in Fig.~\ref{fig:model}. It employs an encoder-decoder architecture. 
The major architecture of $G$ mainly follows the commonly used image-to-image translation networks~\cite{pix2pix,CycleGAN,StarGAN}, which first encodes the input image by a stride of 4, and then decodes it to its original size. 
To improve synthesis realism and diversity for defect generation, we specifically design spatial and categorical control, stochastic variation and layer-based composition for $G$.
The network architecture of $D$ is the same as StarGAN~\cite{StarGAN}, which includes a $D_{src}$ to distinguish faked samples from real ones using PatchGAN~\cite{patchgan} and a $D_{cls}$ to predict the categories of generated defects.

\medskip
\noindent {\bf Spatial and Categorical Control for Defect Generation.}
Different types of defects can exist on different locations of normal samples. To provide better attribute (spatial and categorical) control over the generated defects, we feed an attribute controlling map $A \in R^{H \times W \times C}$ into $G$ to add specific kind of defects to specific location, where $A_{x,y} \in R^C$ represents the presence of defects at the corresponding location, and $C$ denotes the number of defect categories.
$A$ is imposed into the network via SPADE normalization~\cite{SPADE} and is fed into every block in the decoder part of $G$.

Note that since we only assume image-level annotations available, during training stage, the attribute controlling map $A$ should be constant for all locations of the image, i.e., $A$ is acquired by spatial-wisely repeating the target defect label $c \in R^C$. This restriction can be lifted during inference stage, which enables Defect-GAN to add defects at different location(s) in a context-compatible manner.

\medskip
\noindent {\bf Stochastic Variation of Defects.}
Unlike general objects, defects are known to possess complex and irregular patterns with extremely high stochastic variations that are extremely challenging to model using GANs. To mitigate this issue, we employ an adaptive noise insertion module in each block of the encoder-decoder architecture, which explicitly injects Gaussian noise into the feature maps after each convolutional block. For each noise injection, it learns a exclusive scalar to adjust the intensity of the injected noise. By explicitly mirroring the stochastic variations within defects, Defect-GAN can generate more realistic defect samples with much higher diversity.

\medskip
\noindent {\bf Layer-Wise Composition.}
As illustrated in Fig.~\ref{fig:model}, Defect-GAN is also different from existing image-to-image translation GANs~\cite{pix2pix,CycleGAN,StarGAN,SDGAN} in that we consider the final generation as composition of two layers. Specifically, in the defacement process, final defects samples are generated by adding a defect foreground layer on top of the provided normal samples. Similarly, in the restoration process, final restored normal samples are generated by adding a repaint foreground layer on top of the defect samples.

The defacement process can be formulated as:
\begin{equation}  \label{equ:layerbasedsyn1}　
f_d, m_d = G(n,  A_{n \rightarrow d})
\end{equation}
\begin{equation}  \label{equ:layerbasedsyn2}　
d = n \odot (1 - m_d) + f_d \odot m_d
\end{equation}
\noindent where $\odot$ denotes spatial-wise multiplication, $f_d$ denotes generated defect foreground, and $m_d \in [0,1]$ denotes the corresponding spatial distribution map of $f_d$.
Similarly, the restoration process can be formulated as:
\begin{equation}  \label{equ:layerbasedsyn3}　
f_{\hat{n}}, m_{\hat{n}} = G(d,  A_{d\rightarrow \hat{n}})
\end{equation}
\begin{equation}  \label{equ:layerbasedsyn4}　
\hat{n} = d \odot (1 - m_{\hat{n}}) + f_{\hat{n}} \odot m_{\hat{n}}
\end{equation}
\noindent where $\hat{n}$ denotes restored normal sample without defects.

\smallskip
The intuition behind this layer-wise composition strategy is that defects can be deemed a special kind of foreground composed on the background (normal samples). Similarly, the restoration process that removes defects from the background can also be considered a `repainting' process to cover the defect areas. Instead of generating synthesized images directly, Defect-GAN separately generates defect foregrounds along with the corresponding spatial distribution maps, and then performs an layer-wise composition to produce the synthesized defect samples.

The novel compositional layer-based synthesis can significantly improve defect synthesis in terms of both realism and diversity. 
This is mainly because by taking normal samples as background, our model implicitly focuses on generation of defects, without considering the generation of backgrounds. This feature provides our model with more capability to generate more realistic defect samples. 
Furthermore, defects can potentially exists on various backgrounds. Due to the rarity of defect samples, we can only collect specific defects on a very limited number of backgrounds. As a result, typical image synthesis methods lack defect transferablity, i.e., they can only synthesize defect samples under a constrained number of contexts. 
Our proposed layer-wise composition strategy can mitigate this issue. This is because it is able to sufficiently preserve the identities (appearances, styles, etc.) of backgrounds, which forces the model to simulate how defects would interact with the exact provided backgrounds. This significantly improves the defect transferability, which means our model is capable of generating new defect samples within variously different backgrounds.

\medskip
\noindent {\bf Training Objective.}
To generate visually realistic images, we adopt adversarial loss to make the generated defect $d$ indistinguishable from real defect sample $d_{real}$.
\begin{align}\begin{aligned} \label{equ:advloss}
\mathcal{L}_{adv} = \mathop{min}\limits_{G}\mathop{max}\limits_{D_{src}}\mathbb{E}_{d_{real}}[logD_{src}(d_{real})] \\
+ \mathbb{E}_{d}[log(1-D_{src}(d))]
\end{aligned}\end{align}

Our Defect-GAN aims to generate defects conditioned on target defect label $c \in R^C$. To make the generated defects align with the target category, we impose a category classification loss, which consists of two terms: $\mathcal{L}_{cls}^{r} $ to optimize $D$ by classifying real defect sample ${d}_{real}$ to the corresponding category $c'$, and $\mathcal{L}_{cls}^{f}$ to optimize $G$ to generate defect sample of target category $c$. 
\begin{align}\begin{aligned}
 \mathcal{L}_{cls}^{r} = \mathbb{E}_{{d}_{real}, c'}[-log(D_{cls}(c'|d_{real}))]
\end{aligned}\end{align}
\begin{align}\begin{aligned}
\mathcal{L}_{cls}^{f} = \mathbb{E}_{{d}, c}[-log(D_{cls}(c|d))]
\end{aligned}\end{align}

Additionally, we impose a reconstruction loss $\mathcal{L}_{rec}$ that helps preserve the content of input images as much as possible. We adopt L1 loss for the reconstruction loss. 
\begin{align}\begin{aligned}
\mathcal{L}_{rec} = \mathbb{E}_{n, \hat{n}}[||n - \hat{n}||_{1}]
\end{aligned}\end{align}

The layer-wise composition strategy will generate spatial distribution maps in both defacement and restoration process to guide the final compositions.
We further improve composition by introducing two additional spatial constraints (beyond spatial distribution maps), namely, a cycle-consistency loss and a region constrain loss.

To precisely restore the generated defect samples to normal samples, the repaint spatial distribution map shall be ideally the same as the defect spatial distribution map. Thus, we design a spatial distribution cycle-consistency loss $\mathcal{L}_{sd-cyc} $ between the defect spatial distribution map and the repaint spatial distribution map.
\begin{align}\begin{aligned}
\mathcal{L}_{sd-cyc} = \mathbb{E}_{m_n, m_{\hat{n}}}[||m_n - m_{\hat{n}}||_{1}]
\end{aligned}\end{align}

Meanwhile, to avoid the defect foreground and the repaint foreground to take over the whole image area, we introduce a region constrain loss $\mathcal{L}_{sd-con}$ to penalize excessively large defect and foreground distribution maps:
\begin{align}\begin{aligned}
\mathcal{L}_{sd-con} = \mathbb{E}_{m_n, m_{\hat{n}}}[||m_n - 0||_{1} + || m_{\hat{n}}-0||_{1}]
\end{aligned}\end{align}

The overall training objectives for G and D are: 
\begin{align}\begin{aligned}
\mathcal{L}_{D} = -\mathcal{L}_{adv} + \lambda_{cls}^{r}\mathcal{L}_{cls}^{r}
\end{aligned}\end{align}
\begin{align}\begin{aligned}
\mathcal{L}_{G} = \mathcal{L}_{adv} + \lambda_{cls}^{f}\mathcal{L}_{cls}^{f} + \lambda_{rec}\mathcal{L}_{rec} \\
+ \lambda_{con}\mathcal{L}_{sd-cyc} + \lambda_{c}\mathcal{L}_{sd-con}
\end{aligned}\end{align}
where $\lambda_{cls}^{r}$, $\lambda_{cls}^{f}$, $\lambda_{rec}$, $\lambda_{sd-cyc}$, $\lambda_{sd-con}$ are hyper-parameters that are empirically set as $2.0$, $5.0$, $5.0$, $5.0$, $1.0$, respectively.

\subsection{Boosting Defect Inspection Performance}

\begin{figure}[t!] 
\begin{center}
   \includegraphics[width=1.0\linewidth, height=42mm]{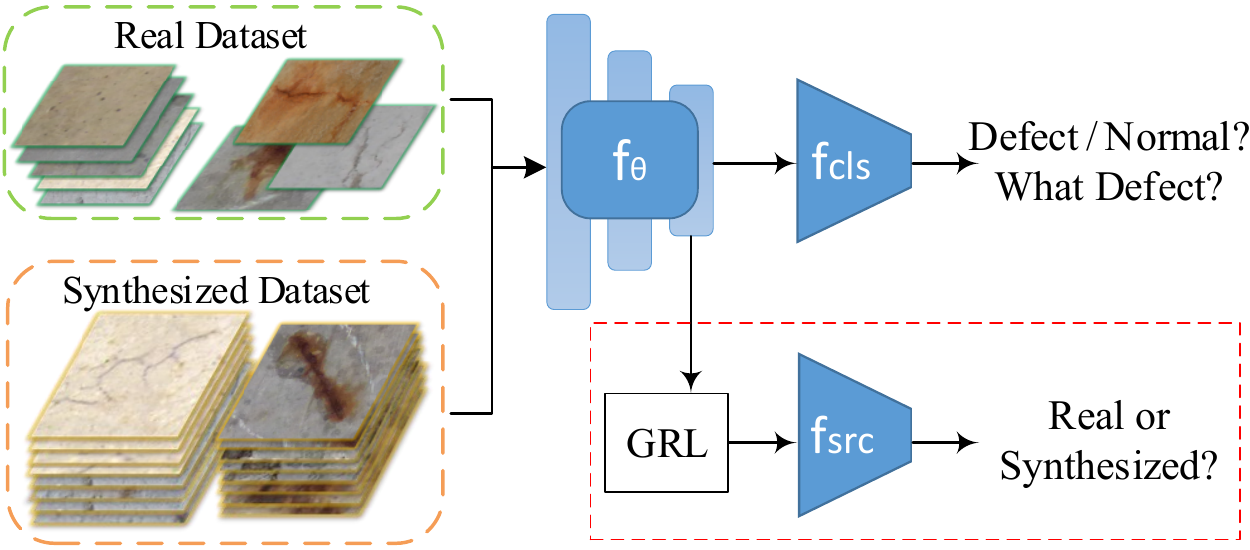}
\end{center}
   \caption{We introduce a source classifier $f_{src}$ (connected to the network backbone $f_\theta$ through a Gradient Reversal Layer) for explicitly distinguishing synthesized and real samples. With this the defect inspection network will not learn for such task undesirably.
   }
\label{fig:recognition}
\end{figure}

The large amounts of defect samples generated by the aforementioned Defect-GAN can be further used to train the state-of-the-art visual recognition models for defect inspection. We adopt the most commonly used image recognition models ResNet~\cite{resnet} and DenseNet~\cite{densenet} to perform defect inspection. The generated defect samples are mixed with the original dataset to train the recognition models.

However, we notice that although Defect-GAN can synthesize realistic defect samples, there still exists a domain gap between the generated samples and the original samples. Naively training a recognition model over the augmented data will lead the model to learn to distinguish these two domains undesirably. We attach an additional source classifier $f_{src}$ to distinguish synthesized samples from real ones explicitly, and connect this domain classifier to the network backbone through a Gradient Reversal Layer (GRL)~\cite{GRL} as illustrated in Fig.\ref{fig:recognition}. Therefore, there will be no distinguishable difference between the features extracted by $f_\theta$ for the synthesized samples and the real samples, which ensures all training data are effectively learnt.

\section{Experiments}

This section presents experimentation of our methods. We first evaluate Defect-GAN's defect synthesis performance, and then demonstrate its capacity in boosting defect inspection performance as a data augmentation method.

\smallskip
\noindent {\bf Dataset.}
We evaluate Defect-GAN on CODEBRIM\footnote{Dataset available at https://doi.org/10.5281/zenodo.2620293}~\cite{CODEBRIM} -- a defect inspection dataset in context of concrete bridges, which features six mutually non-exclusive classes: crack, spallation, efflorescence, exposed bars, corrosion and normal samples. 
%Its classification dataset contains 5,244 defect image patches and 2,485 normal image patches, which are cropped from 1,590 original full-resolution images. 
It provides image patches for multi-label classification as well as the full-resolution images from which image patches are cropped.
Compared with existing open datasets for defect inspection~\cite{CrackForest,CSSC,SDNET2018}, CODEBRIM is the most challenging and complex one to the best of our knowledge, which can better reflect the practical scenarios.

\begin{table}
\begin{center}
\begin{tabular}{c c}
\toprule
\rule{0pt}{14pt} Methods & FID Scores $\downarrow$ \\
\midrule \midrule
StackGAN++~\cite{stackgan++} & 111.1  \\
Conditional StackGAN++~\cite{stackgan++} & 132.1  \\
StyleGAN v2~\cite{StyleGANv2} & 148.2 \\ 
StyleGAN v2~\cite{StyleGANv2} + DiffAug~\cite{diffaugment} & 142.4 \\ 
\hline
CycleGAN~\cite{CycleGAN} & 94.5\\
StarGAN~\cite{StarGAN} & 295.1  \\
StarGAN~\cite{StarGAN} + SPADE~\cite{SPADE} & 103.0  \\
Defect-GAN (Ours) & \bf{65.6}  \\ 
\hline
Ideal Defect Synthesizer & 25.0 \\
\bottomrule
\end{tabular}
\end{center}
\caption{Quantitative comparison of Defect-GAN with existing image synthesis methods in Fréchet Inception Distance (FID).}
\label{table:fid}
\end{table}

\subsection{Defect Synthesis}

\noindent {\bf Implementation Details.}
We use all images from the classification dataset to train Defect-GAN. Besides, we collect extra 50,000 normal image patches by simply cropping from the original full-resolution images. All images are resized to $128\times128$ for training. To stabilize the training and generate better images, we replace Eq.~\ref{equ:advloss} with Wasserstein GAN objective with gradient penalty~\cite{wgan,wgan-gp} and perform one generator update every five discriminator updates. We use Adam optimizer~\cite{AdamOpt} with $\beta_1$ = $0.5$ and $\beta_2$ = $0.999$ to train Defect-GAN with the learning rate starting from $2 \times 10^{-4}$ and reducing to $1 \times 10^{-6}$. We set batch size at 4 and the total training iteration at 500,000. The training takes about one day on a single NVIDIA 2080Ti GPU.

\smallskip
\noindent {\bf Evaluation Metric.}
We adopt the commonly used Fréchet Inception Distance (FID)~\cite{fid} to evaluate the realism of synthesized defect samples. Lower FID scores indicate better synthesis realism.

\smallskip
\noindent {\bf Quantitative Experimental Results.}
Table~\ref{table:fid} show quantitative experimental results regarding defect synthesis fidelity, in which the first block includes direct synthesis methods (image synthesis from a randomly sampled latent code), and the second block includes image-to-image translation methods. We also present the FID score of a perfect defect synthesizer in the third block by randomly separating the real defect samples into two sets and computing the FID score between them.
As Table~\ref{table:fid} shows, the direct synthesis methods generally have unsatisfactory performances due to the lack of defect training samples as well as their limited capacities to capture the complex and irregular patterns of defects. 
As a comparison, by mimicking the defacement and restoration processes following Defect-GAN, existing image-to-image translation methods can generate defects with significantly better quality. This is because, with more information as input, such methods are generally more data-efficient. Besides, they can utilize the large amount of normal samples in training. 
On the other hand, Defect-GAN achieves significantly better synthesis FID, which demonstrate its superiority in defect synthesis. Interestingly, models with categorical control tend to perform worse in terms of FID scores than models without. We believe introducing additional categorical control can limit model's synthesis realism. However, even with such constraint, Defect-GAN still achieves the best performance.

\begin{table}
\begin{center}
\begin{tabular}{ c | c | c | c | c }
\toprule
\multicolumn{4}{c|}{Design Choices} & \multirow{2}{*}{FID Scores $\downarrow$} \\
 \cline{1-4}
 SCC & ANI & LWC & SC & \\
\midrule
$\times$ & $\times$ & $\times$ & $\times$ & 295.1 \\
$\checkmark$ & $\times$ & $\times$ & $\times$ & 103.0 \\
$\checkmark$ & $\checkmark$ & $\times$ & $\times$ & 99.7 \\
$\checkmark$ & $\times$ & $\checkmark$ & $\times$ & 76.8 \\
$\checkmark$ & $\times$ & $\checkmark$ & $\checkmark$ & 69.5 \\
$\checkmark$ & $\checkmark$ & $\checkmark$ & $\checkmark$ & {\bf 65.6} \\
\bottomrule
\end{tabular}
\end{center}
\caption{Ablation studies of the proposed Defect-GAN: Our designed Spatial and Categorical Control (SCC), Adaptive Noise injection (ANI), Layer-Wise Composition (LWC), and additional Spatial Constraints (SC) are complementary and jointly beneficial to the quality of the synthesized defects.}
\label{table:ablation}
\end{table}

We further demonstrate the effectiveness of our proposed designs in Defect-GAN by presenting quantitative ablative experiments in Table~\ref{table:ablation}. Without our designed components, Defect-GAN degrades to StarGAN~\cite{StarGAN} -- a widely used multi-domain image-to-image translation model. However, it fails to converge on this task and cannot synthesize any defect-like patterns. By incorporating Spatial and Categorical Control (SCC), it can converge and generate defect samples with comparable quality with existing methods. Based on this, the Layer-Wise Composition (LWC) can significantly improve the synthesis realism. We believe the reason is twofold: (1) it lifts the defect-insufficiency constraint by allowing the networks to fully focus on defect generation; (2) it can generate contextually more natural defects. Furthermore, Adaptive Noise Injection (ANI) and additional Spatial Constraints (SC) for training can also boost defect synthesis performance. These proposed components are proved to be complementary to each other, enabling Defect-GAN to achieve state-of-the-art defect synthesis quality.

\begin{figure*}[t!] 
\begin{center}
   \includegraphics[width=0.965\linewidth]{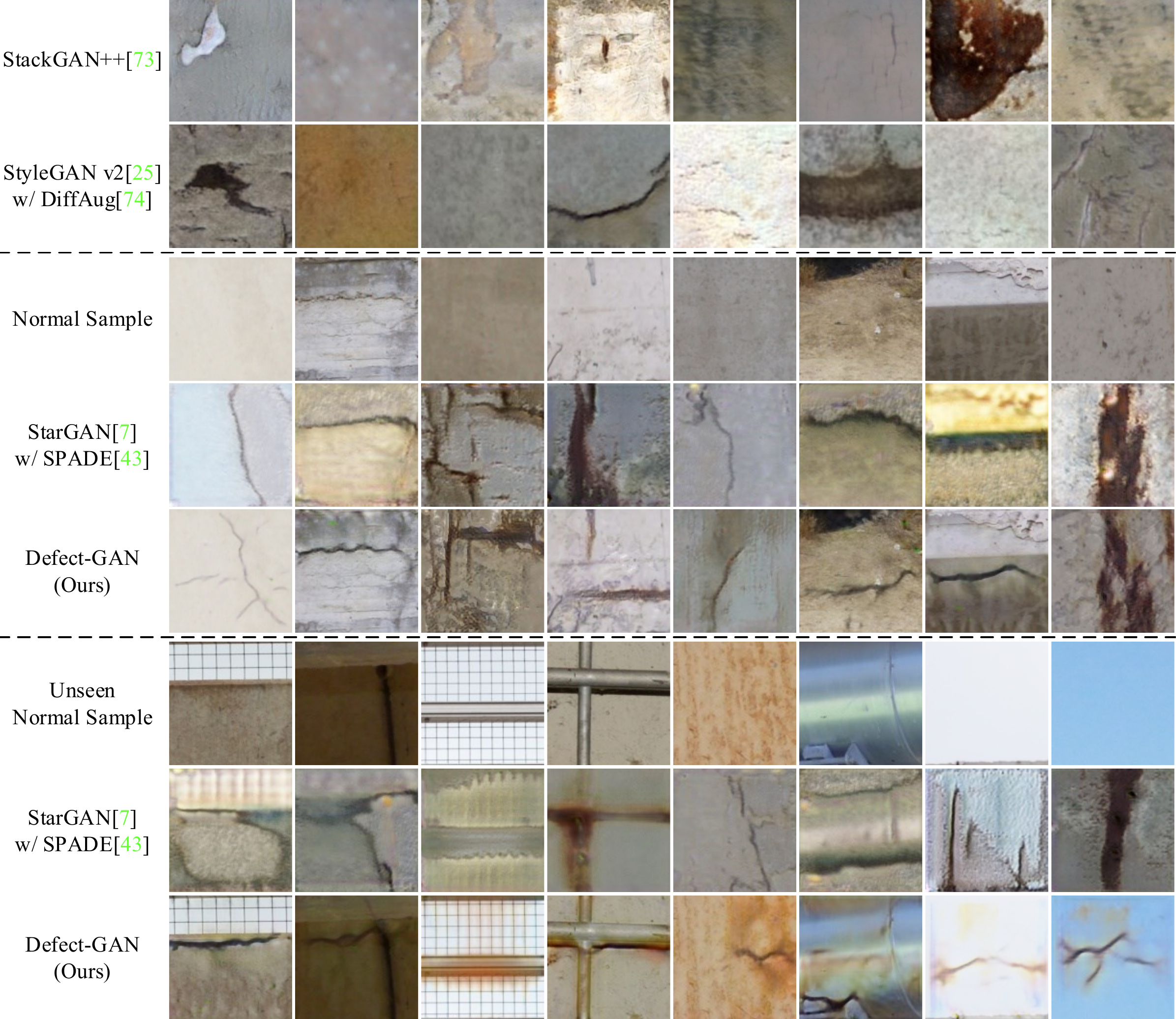}
\end{center}
   \caption{Qualitative comparison of Defect-GAN with state-of-the-art image synthesis methods: Rows 1-2 show direct defect synthesis from random noises by two latest image synthesis methods. Rows 4-5 compare defect generation over \textit{Normal Samples} in Row 3 (used in network training) by StarGAN with SPADE and our Defect-GAN, while Rows 7-8 compare defect generation over \textit{Unseen Normal Samples} in Row 6 (not used in network training) by StarGAN with SPADE and our Defect-GAN.
   }
\label{fig:compare}
\end{figure*}

\begin{figure}[t!] 
\begin{center}
   \includegraphics[width=1.0\linewidth]{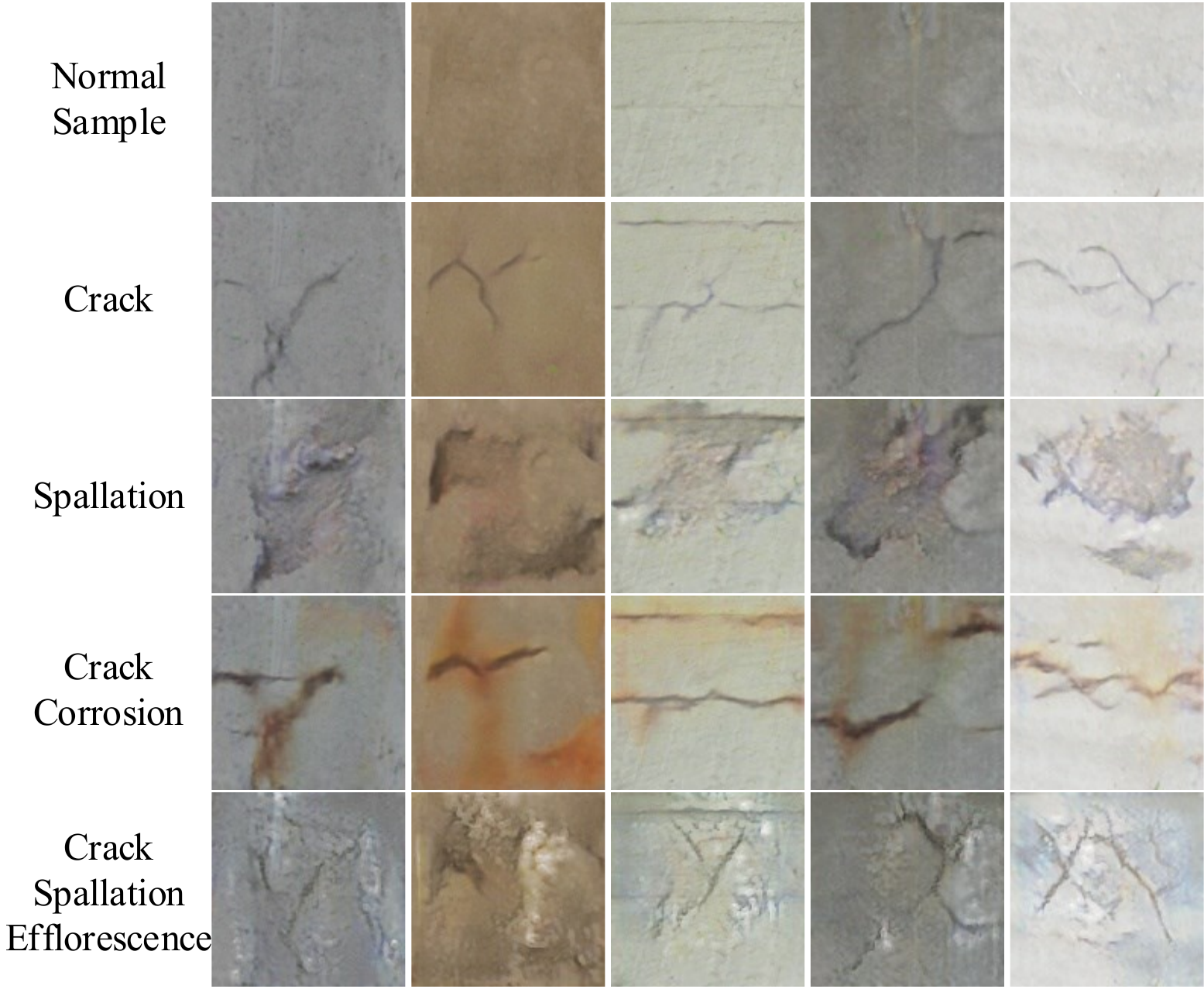}
\end{center}
   \caption{Illustration of categorical control in defect generation by Defect-GAN: For each normal sample in Row 1, Rows 2-3 and 4-5 show the generated defect samples conditioned on a single and multiple target categories, respectively.
   }
\label{fig:category}
\end{figure}

\begin{figure}[t!] 
\begin{center}
   \includegraphics[width=1.0\linewidth]{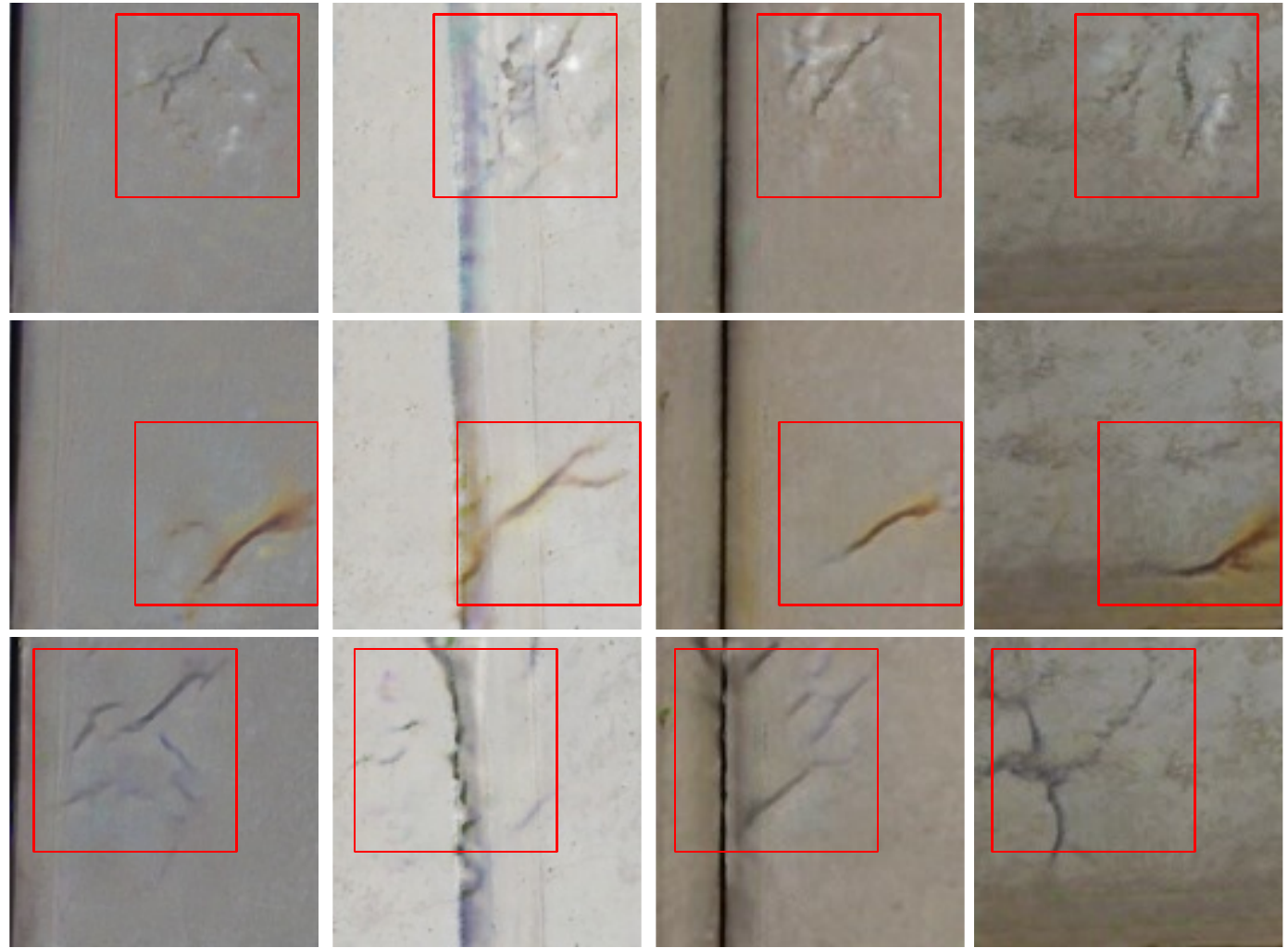}
\end{center}
   \caption{Illustration of spatial control in defect generation by Defect-GAN: Each row shows defect samples generated with different normal samples but the same spatial control, while each column shows defect samples generated with the same normal sample but different spatial controls.
   }
\label{fig:spatial}
\end{figure}

\medskip
\noindent {\bf Qualitative Experimental Results.}
Fig.~\ref{fig:compare} shows qualitative results of Defect-GAN and comparisons with other synthesis methods. Rows 1-2 show the synthesis by state-of-the-art direct synthesis methods: StackGAN++~\cite{stackgan++} and StyleGAN v2~\cite{StyleGANv2} with DiffAug~\cite{diffaugment}. We can see that many generated samples do not contain clear defects, and some samples are not visually natural. This verifies the aforementioned limitation of GANs for defect synthesis.
For image-to-image translation methods, we choose StarGAN~\cite{StarGAN} with SPADE~\cite{SPADE} as the competing method since it offers categorical control as Defect-GAN. And other methods like CycleGAN~\cite{CycleGAN} and SDGAN~\cite{SDGAN} produce visually similar results.
As shown in Row 4-5, StarGAN w/ SPADE and Defect-GAN can produce visually realistic and diverse defect samples conditioned on normal samples. Defect samples by StarGAN w/ SPADE look comparable with Defect-GAN, except that it tends to alter the background identity, while Defect-GAN can preserve the appearance and style of normal samples thanks to the layer-wise composition strategy.
On the other hand, StarGAN w/ SPADE completely fails to transfer the learnt defect patterns to novel backgrounds that are not seen during training, while Defect-GAN shows superb defect transferability and synthesis realism as shown in Rows 7-8. This property is essential for introducing new information into the training data.

In addition, we show Defect-GAN's categorical control in defect generation in Fig.~\ref{fig:category}, where different types of defects can be generated conditioned on the same normal image. Fig.~\ref{fig:spatial} also shows Defect-GAN's spatial control in defect generation, where red boxes denote the intended places to generate defects. Defect-GAN can generate defects on specific locations while maintaining contexts natural.

\subsection{Defect Inspection}

\noindent {\bf Implementation Details.}
We use the training set and additional 50,000 normal images to train Defect-GAN. Then, Defect-GAN expands the training samples by synthesizing 50,000 defect samples. The generated defect samples are mixed with the original training data to train the defect inspection networks, with the extra restored normal samples also included to avoid data imbalance. All images are resized to $224\times224$. We adopt SGD with a learning rate of $1\times10^{-3}$ to train the network until convergence. Batch size is set to $16$. We use the validation set to select the best performing model and report the performance on the test set.

\smallskip 
\noindent {\bf Quantitative Experimental Results.}
As CODEBRIM features a multi-label classification task, we can only adopt methods with categorical control to expand the training samples.
Results for defect inspection is shown in Table~\ref{table:defectinspection}. The first row of each block shows defect inspection performance only with original training data, and the rest three rows of each block present defect inspection performance with original training data and the augmented samples generated by different synthesis methods. For fair comparison, 50,000 synthesised defect samples are augmented for all synthesis methods.
As the results shows, the synthesized defect samples from Conditional StackGAN++~\cite{stackgan++} greatly drop the defect inspection performance. This is because that StackGAN++ is not even able to generate realistic defect samples due to its limited capacity in defect modeling. StackGAN++ generated defect samples are harmful to network training.
On the other hand, StarGAN\cite{StarGAN}+SPADE\cite{SPADE} generated samples can slightly boost the inspection performance. And our proposed Defect-GAN can further significantly improve the accuracy of trained defect inspection networks. Although both methods can generate defect samples with good visual realism, our proposed Defect-GAN is capable of simulating the learnt defects on backgrounds that are not seen during training. This feature makes Defect-GAN generated samples much more diverse, thus can introducing new information into the training data, significantly improving the performance of trained models. The results also demonstrate the superiority of Defect-GAN for defect synthesis in terms of fidelity, diversity and transferability.

\begin{table} \small
\begin{center}
\resizebox{0.5\textwidth}{!}{
\begin{tabular}{c c c }
\toprule
Networks & Augmentation Methods & Accuracy(\%)\\
\midrule \midrule
% Texture-CNN~\cite{Texture-CNN} & 67.93 \\
% ENAS~\cite{enas} & 70.78 \\
% Meta-QNN~\cite{metaqnn} & 72.19 \\
% \midrule
\multirow{4}{*}{ ResNet34~\cite{resnet}}& None & 70.25 \\
& Conditional StackGAN++\cite{stackgan++}   & 62.59 \\
& StarGAN\cite{StarGAN}+SPADE\cite{SPADE} & 71.90 \\
& Defect-GAN (Ours) & \bf{75.48} \\
\midrule
\multirow{4}{*}{DenseNet121~\cite{densenet}}& None & 70.77 \\
& Conditional StackGAN++\cite{stackgan++}   & 58.68 \\
& StarGAN\cite{StarGAN}+SPADE\cite{SPADE} & 72.61 \\
& Defect-GAN (Ours) & \bf{75.79} \\
\bottomrule
\end{tabular}
}
\end{center}
\caption{Quantitative experimental results for defect inspection.}
\label{table:defectinspection}
\end{table}

\section{Conclusion}
This paper presents a novel Defect-GAN for defect sample generation by mimicking the defacement and restoration processes. It can capture the stochastic variations within defects and can offer flexible control over the locations and categories of the generated defects. Furthermore, with a novel compositional layer-based architecture, it is able to generate defects while preserving the style and appearance of the provided backgrounds. The proposed Defect-GAN is capable of generating defect samples with superior fidelity and diversity, which can further significantly boost the performances of defect inspection networks.

\medskip
\medskip
\smallskip
\smallskip
\noindent
\textbf{Acknowledgments:}
This work was conducted within the Delta-NTU Corporate Lab for Cyber-Physical Systems with funding support from Delta Electronics Inc. and the National Research Foundation (NRF) Singapore under the Corp Lab @ University Scheme (Project No.: DELTA-NTU CORP-SMA-RP15).

\newpage
{\small
\bibliographystyle{ieee_fullname}
\bibliography{egbib}
}

\clearpage

\begin{figure*}[!t] 
\begin{center}
   \includegraphics[width=1.0\linewidth]{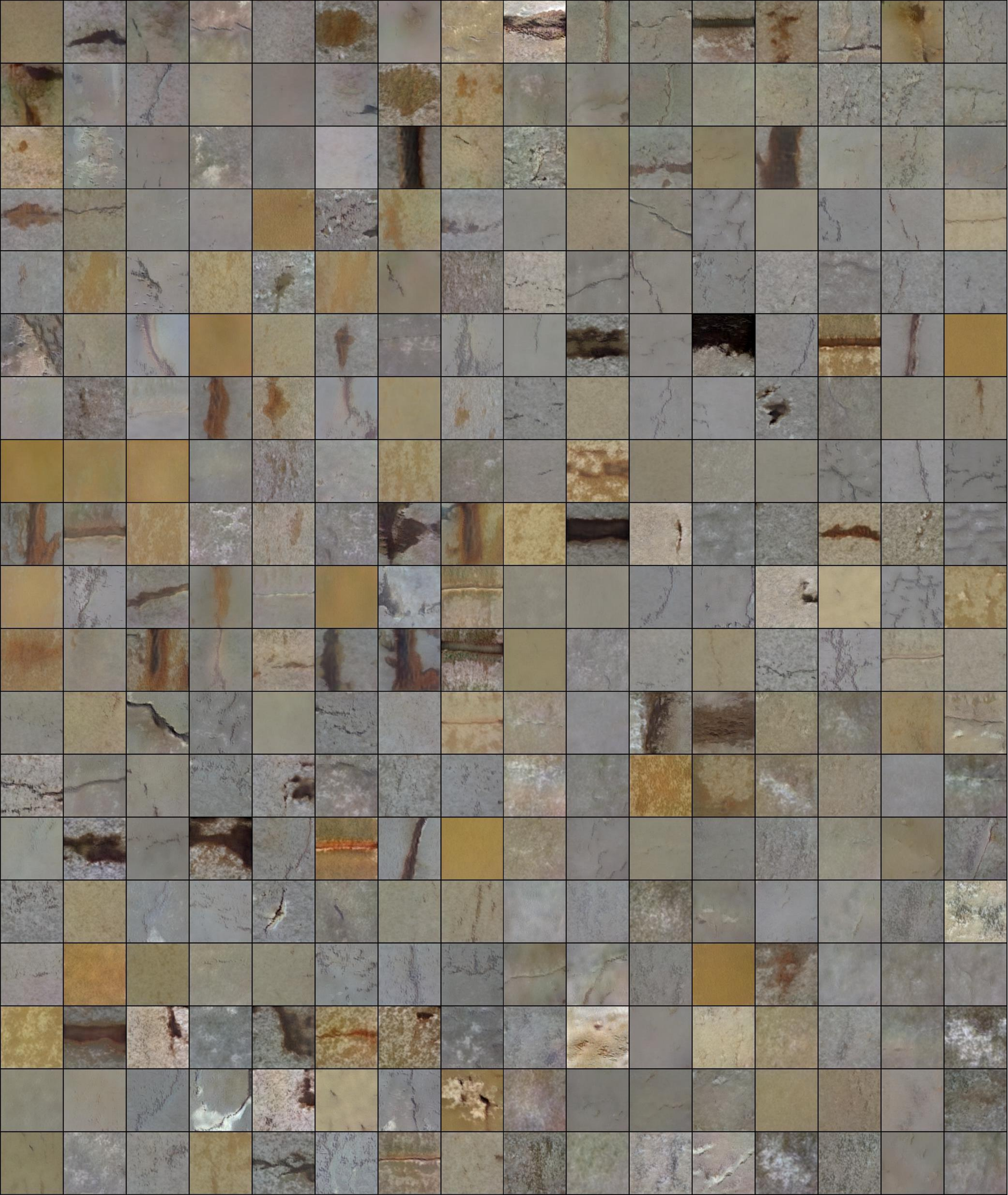}
\end{center}
   \caption{Randomly sampled defect samples generated by StyleGAN v2 w/ DiffAug: Although it is able to generate some visually realistic defect samples, a lot of the generated samples still do not contain any defects, which verifies the limitation in model's capacity to capture and model the complex and irregular textures of defects.
   }
\label{styleGAN}
\end{figure*}

\begin{figure*}[!t] 
\begin{center}
   \includegraphics[width=1.0\linewidth]{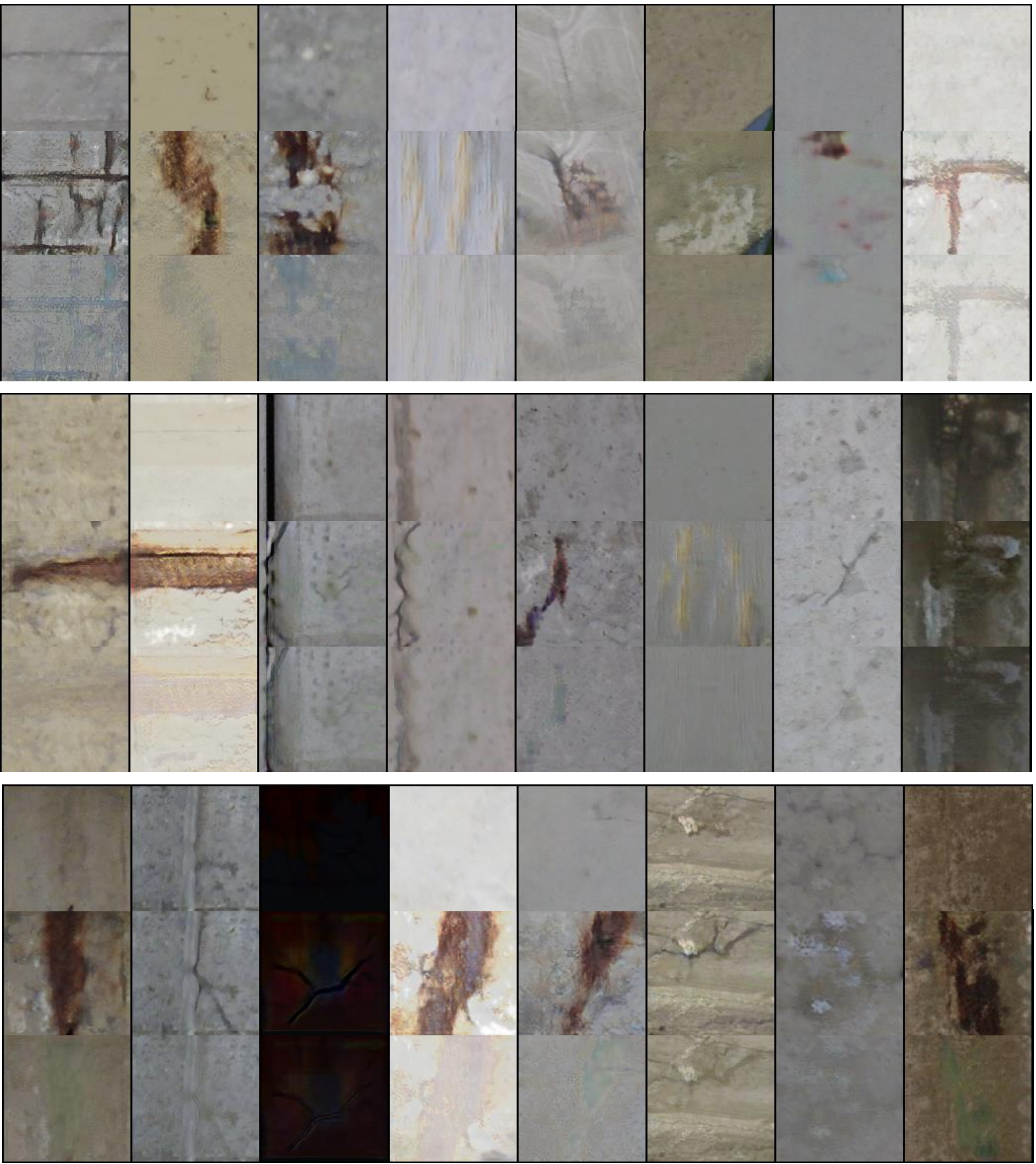}
\end{center}
   \caption{Additional results of the proposed Defect-GAN to mimick the defacement and restoration processes: For each block, the first row contains real normal samples; the second and third row contain generated defect samples and restored normal samples, respectively.
   }
\label{Defect-GAN}
\end{figure*}

\begin{figure*}[!t] 
\begin{center}
   \includegraphics[width=1.0\linewidth]{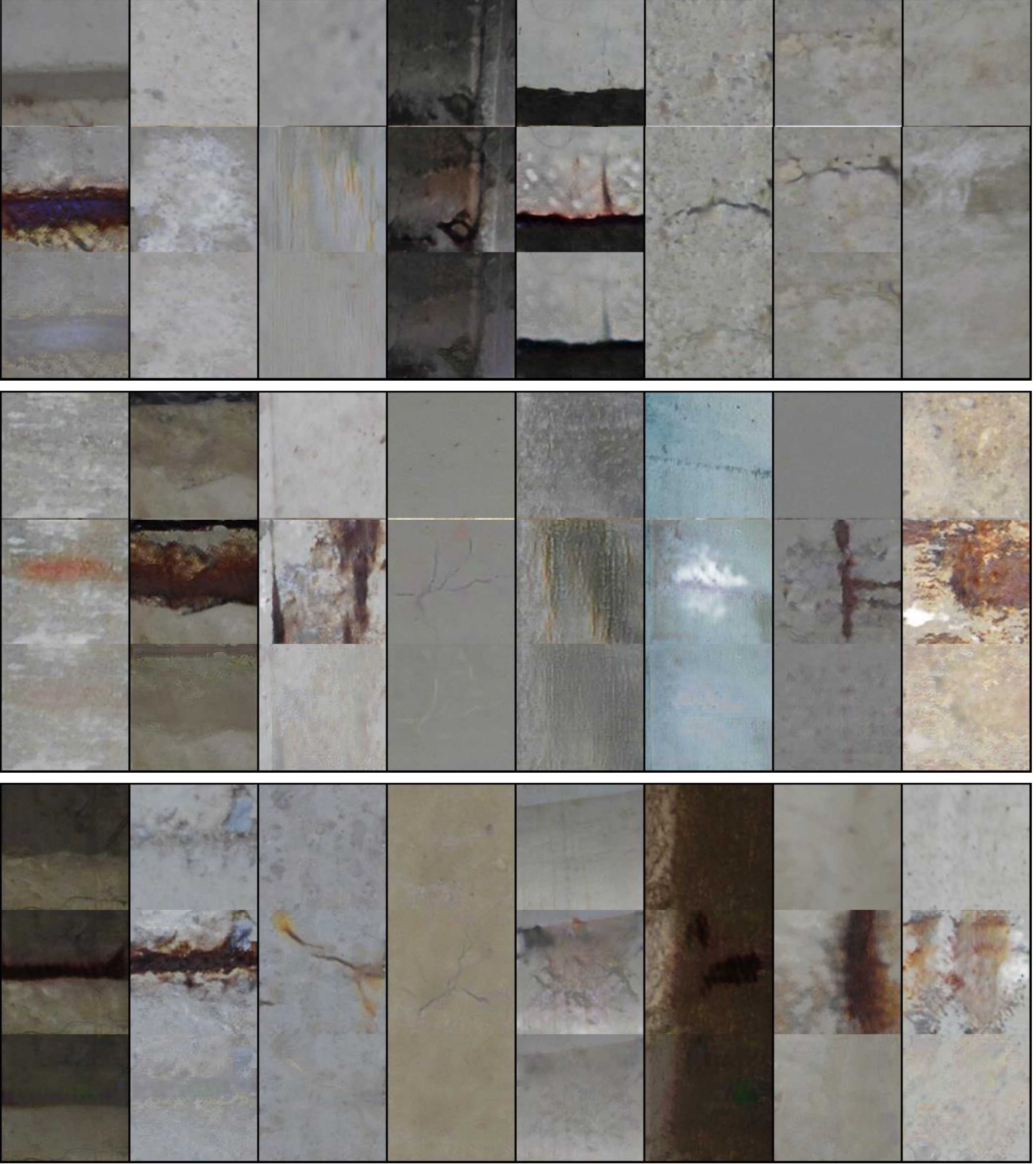}
\end{center}
   \caption{Additional results of the proposed Defect-GAN to mimick the defacement and restoration processes: For each block, the first row contains real normal samples; the second and third row contain generated defect samples and restored normal samples, respectively.
   }
\label{Defect-GAN1}
\end{figure*}

\begin{figure*}[!t] 
\begin{center}
   \includegraphics[width=1.0\linewidth]{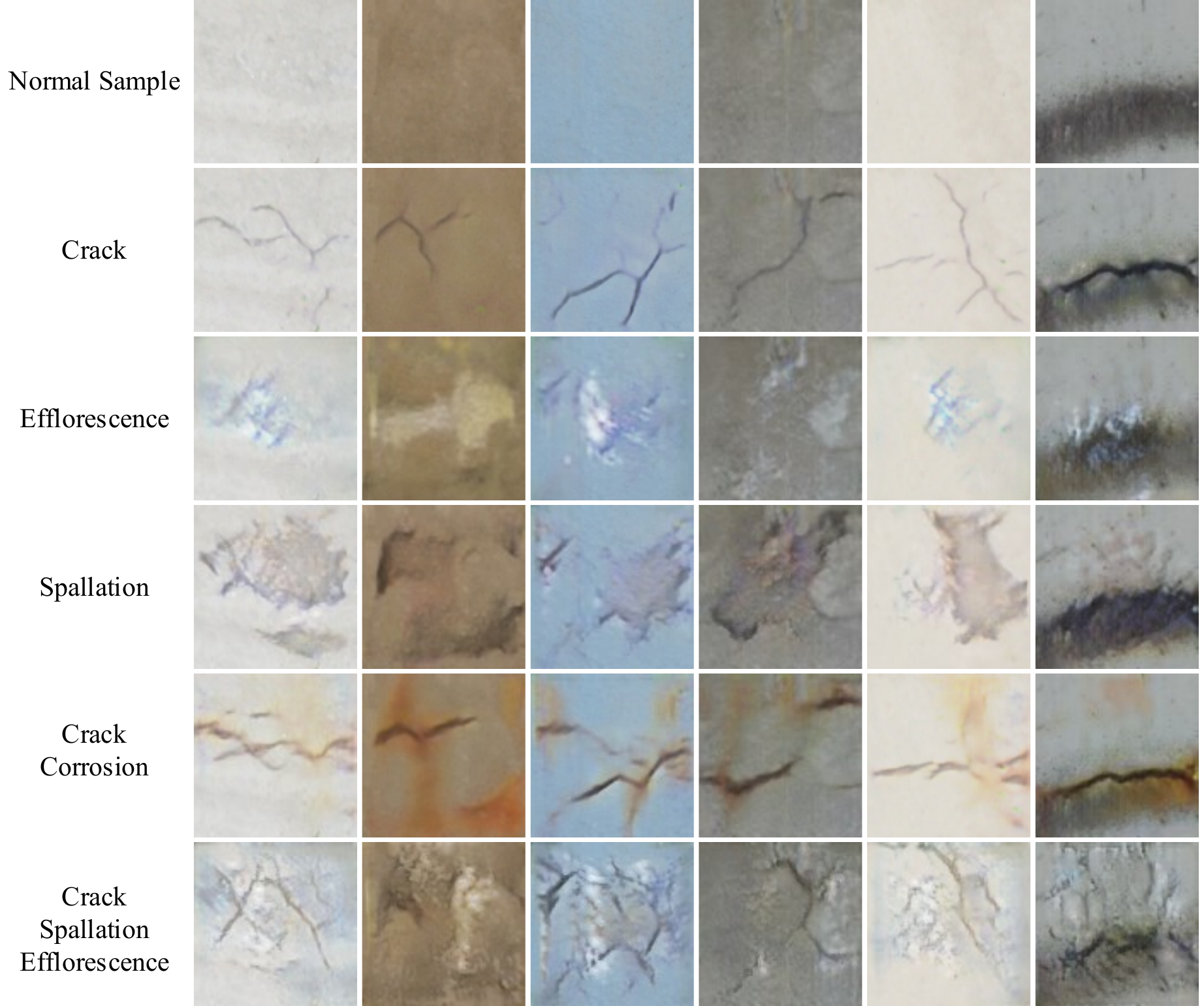}
\end{center}
   \caption{Illustration of categorical control in defect generation by Defect-GAN: For each normal sample in Row 1, Rows 2-6 show the generated defect samples conditioned on target categories, respectively.
   }
\label{category}
\end{figure*}

\begin{figure*}[!t] 
\begin{center}
   \includegraphics[width=1.0\linewidth]{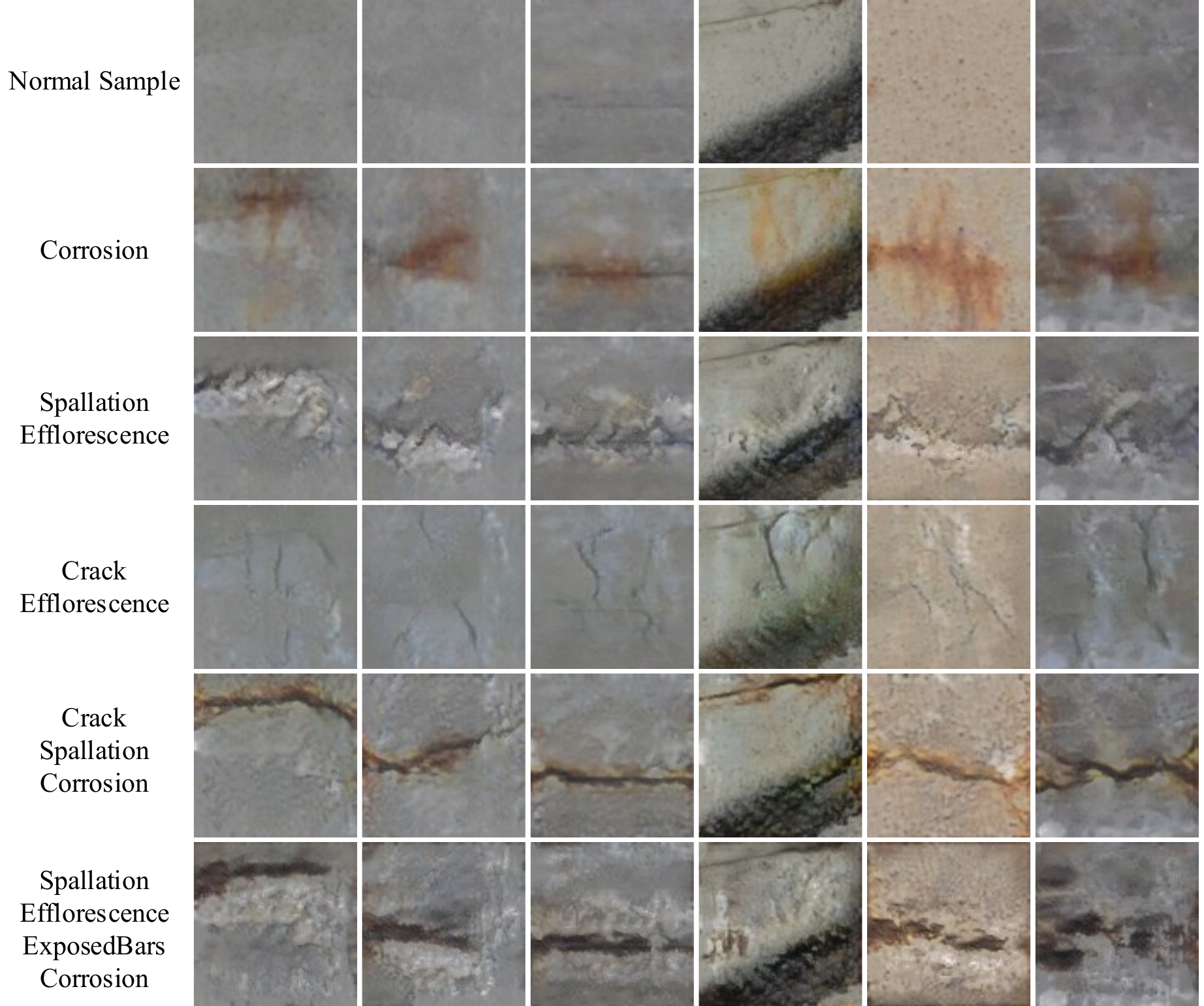}
\end{center}
   \caption{Illustration of categorical control in defect generation by Defect-GAN: For each normal sample in Row 1, Rows 2-6 show the generated defect samples conditioned on target categories, respectively.
   }
\label{category1}
\end{figure*}

\end{document}